\documentclass[sigconf,dvipsnames]{acmart}

\acmConference[GECCO '26]{Genetic and Evolutionary Computation Conference}{July 2026}{San José, Costa Rica}
\acmBooktitle{Proceedings of the Genetic and Evolutionary Computation Conference (GECCO '26), July 2026, San José, Costa Rica}

\renewcommand\footnotetextcopyrightpermission[1]{}
\acmDOI{}
\acmISBN{}

\acmBooktitle{}

\AtBeginDocument{%
  }

\usepackage{graphicx}
\usepackage{booktabs}
\usepackage{amsmath,amsfonts}
\usepackage{amssymb}
\usepackage{algorithm}
\usepackage{algpseudocode}
\usepackage{enumitem}
\usepackage{appendix}
\usepackage{subfig}

\title{Population Dynamics in ARIEL Robotics Systems Featuring Embodied Evolution via Spatial Mating Mechanisms}
\author{Victoria Peterson}
\affiliation{%
  \institution{Vrije Universiteit}
  \city{Amsterdam}
  \country{Netherlands}
}

\author{Akshat Srivastava}
\affiliation{%
  \institution{Vrije Universiteit}
  \city{Amsterdam}
  \country{Netherlands}
}

\author{Raghav Prabhakar}
\affiliation{%
  \institution{Vrije Universiteit}
  \city{Amsterdam}
  \country{Netherlands}
}

\begin{abstract}
We present a Spatially Embedded Evolutionary Algorithm where robot individuals exist in a physically simulated 2D environment, must navigate to encounter potential mates, and compete for survival under various spatially-aware selection pressures. Using HyperNEAT evolved neural controllers for ARIEL gecko-inspired quadrupeds in MuJoCo, we investigate how spatial structure fundamentally alters evolutionary dynamics. Our experiments show a modest 4.9\% difference in peak fitness between proximity-based and random pairing---possibly within stochastic variation---while combining spatial parent selection with stochastic death selection produces unstable population dynamics. We discover a continuous phase transition in energy-based selection experiments, with critical zone count $n_c \approx 14.7$ separating extinction-dominated and explosion-dominated regimes. Our density-dependent death selection mechanism achieves 97\% completion rates but causes fitness decline, revealing a fundamental dilemma where decoupled mechanisms produce bistable dynamics, positively coupled mechanisms create counter-selection pressures, and only deterministic fitness-based selection maintains stability. These findings provide important constraints for future spatial EA design.
\end{abstract}



\begin{document}
\maketitle

\section{Introduction}
Evolutionary Algorithms (EAs) have been widely used to evolve robotic controllers and morphologies, typically under abstract evolutionary assumptions such as panmictic populations, where any individual may reproduce with any other. While computationally convenient, these abstractions ignore spatial constraints that are fundamental to biological evolution, where interactions are local and shaped by physical proximity. Such constraints introduce localized selection pressure, genetic drift, and population clustering, all of which influence evolutionary dynamics and emergent behavior.

In contrast, spatially structured evolutionary systems embed individuals in space and restrict interactions to local neighborhoods. This induces \emph{isolation by distance}, slowing the global diffusion of high-fitness genotypes and promoting the emergence of spatial niches. Prior work has shown that these dynamics can preserve diversity and mitigate premature convergence, but also lead to heterogeneous fitness distributions and localized behavioral specialization compared to globally mixing populations \cite{chopard_parallel_2000, pappa_genetic_2023}.

This work investigates spatial evolution in the context of embodied robotics, where evolutionary processes unfold through the physical interactions of autonomous agents rather than centralized evaluation. We present a Spatial Evolutionary Algorithm in which ARIEL robot individuals exist in a physically simulated 2D environment, must navigate to encounter potential mates, and compete for reproduction under spatially mediated selection pressures. We tightly integrate evolutionary computation with physics-based simulation via ARIEL MuJoCo, where spatial dynamics, movement, and resource constraints jointly shape adaptation. In this setting, the environment itself implicitly applies selection pressure, shifting the evolutionary process from objective-driven optimization toward open-ended ecological interaction \cite{doncieux_new_2011, bredeche_environment-driven_2012, bredeche_embodied_2018}.

\subsection*{Research Questions}

Our investigation addresses four key questions at the intersection of spatial evolution and embodied robotics:

\subsubsection*{RQ1: How does spatial structure affect Evolutionary Algorithm Dynamics?}


\subsubsection*{RQ2: How do different selection pressures impact adaptation in Spatial Environments?}


\subsubsection*{RQ3: How can non-spatial Parent Selection Operators be modified to apply spatially?}


\subsubsection*{RQ4: Can spatial Parent Selection and Death Selection operators be designed to balance population growth?}


\section{Related Work}
This research builds on three intersecting lines of work: spatially structured evolutionary algorithms, embodied evolution, and population dynamics in decentralized evolutionary systems.

\subsection{Spatially Structured Evolutionary Algorithms}
To mitigate premature convergence in globally mixing populations, spatially structured EAs keep interactions in local sub-populations, suing approaches such as coarse-grained Island Models, where distinct populations exchange individuals via migration, and fine-grained Cellular EAs, where individuals interact only with immediate neighbors on a grid \cite{kacprzyk_parallel_2015, skolicki_analysis_2005, kroc_simulating_2010}. Spatial restriction slows the spread of advantageous genotypes, maintaining diversity through localized competition and reduced gene flow.

Our work extends these concepts beyond fixed topologies to continuous physical space, where neighborhoods emerge dynamically in flat 2D space through movement and proximity rather than predefined grid or island structures.

\subsection{Embodied Evolution and Interactive Ecosystems}
Embodied Evolution (EE) distributes evolutionary processes across populations of autonomous robots, eliminating centralized evaluation and conducting selection and reproduction online during system operation \cite{doncieux_new_2011, bredeche_embodied_2018}. This paradigm has evolved into Interactive Robot Ecosystems, such as mEDEA, where agents continuously exchange genetic material through local encounters rather than explicit fitness ranking \cite{9774747}.

In these systems, survival and reproductive success depend on an agent’s ability to remain operational within its environment, so selection pressure is implicit and environment-driven, with outcomes affected by navigation ability, energy management, and encounter frequency. Rather than optimizing a predefined objective, adaptation emerges from sustained interaction between agents and their surroundings, making EE good for studying spatial evolutionary dynamics.

\subsection{Mating and Population Dynamics}
Mate selection in decentralized evolutionary systems is opportunistic and local, relying on physical encounters rather than global fitness comparisons, which helps preserve diversity, but can also favor behaviors that maximize encounter rates or clustering over exploration \cite{diependaal_how_nodate}. 

Unlike traditional EAs with fixed population sizes, embodied ecosystems exhibit dynamic populations, while energy is often used as a proxy for fitness, where agents must balance survival costs with reproductive investment. This coupling of energy, movement, and reproduction creates a natural mechanism for population regulation while introducing trade-offs between exploration, mating success, and longevity \cite{yao_parallel_2004, bredeche_environment-driven_2012}.

\section{Theory}
This section outlines the theoretical basis and core mechanisms of our spatial evolutionary system: spatially structured EAs, HyperNEAT control representations, and the parent/death selection operators that govern reproduction and survival in a 2D embodied environment. Table~\ref{tab:notation} summarizes the notation used throughout.

\begin{table}
\centering
\caption{Summary of notation used in this work.}
\label{tab:notation}
\small
\begin{tabular}{@{}lll@{}}
\toprule
\textbf{Symbol} & \textbf{Description} & \textbf{Section} \\
\midrule
\multicolumn{3}{@{}l}{\textit{HyperNEAT / Controller}} \\
$w, w_{ij}$ & Connection weight (from neuron $i$ to $j$) & \ref{sec:hyperneat} \\
$(x_1,y_1,x_2,y_2)$ & Source and target neuron coordinates & \ref{sec:hyperneat} \\
$n$ & Number of robot joints & \ref{sec:hyperneat} \\
$\tau$ & Weight pruning threshold & \ref{sec:hyperneat} \\
\midrule
\multicolumn{3}{@{}l}{\textit{Spatial Structure}} \\
$\mathbf{x}_i$ & Position of individual $i$ in world space & \ref{sec:parent_sel} \\
$d_{ij}$ & Periodic distance between individuals $i$ and $j$ & \ref{sec:parent_sel} \\
$W, H$ & World width and height & \ref{sec:periodic} \\
\midrule
\multicolumn{3}{@{}l}{\textit{Parent Selection}} \\
$r_{\text{pair}}$ & Pairing radius for proximity mating & \ref{sec:parent_sel} \\
$r_{\text{zone}}$ & Radius of mating zones & \ref{sec:parent_sel} \\
$\mathbf{c}_k$ & Center position of mating zone $k$ & \ref{sec:parent_sel} \\
$P$ & Set of already-paired individuals & \ref{sec:parent_sel} \\
$n_{\text{gen}}$ & Generations between zone relocations & \ref{sec:parent_sel} \\
$\mathbf{d}_{\text{input}}$ & Directional input vector to controller & \ref{sec:movement} \\
\midrule
\multicolumn{3}{@{}l}{\textit{Death Selection}} \\
$N$ & Target population size & \ref{sec:death_sel} \\
$\mathrm{age}_i, \mathrm{age}_{\max}$ & Age of individual $i$; maximum age & \ref{sec:death_sel} \\
$E_i$ & Energy level of individual $i$ & \ref{sec:death_sel} \\
$\delta E_{\text{depletion}}$ & Energy lost per generation & \ref{sec:death_sel} \\
$\Delta E_{\text{mating}}$ & Energy change from mating & \ref{sec:death_sel} \\
$\rho_i$ & Local density around individual $i$ & \ref{sec:death_sel} \\
$\rho_c$ & Critical density threshold & \ref{sec:death_sel} \\
$\sigma$ & Locality radius (density kernel width) & \ref{sec:death_sel} \\
$P_{\text{base}}, P_{\text{max}}$ & Base and max death probabilities & \ref{sec:death_sel} \\
\midrule
\multicolumn{3}{@{}l}{\textit{Fitness Evaluation}} \\
$f_{\text{simple}}, f_{\text{directional}}$ & Fitness functions & \ref{sec:fitness} \\
$\mathbf{x}_{\text{start}}, \mathbf{x}_{\text{end}}$ & Start and end positions & \ref{sec:fitness} \\
$d_{\text{total}}$ & Total distance traveled & \ref{sec:fitness} \\
$w_{\text{progress}}$ & Weight for directional bonus & \ref{sec:fitness} \\
$b_{\text{direction}}$ & Directional progress score & \ref{sec:fitness} \\
$\mathbf{u}_{\text{target}}$ & Unit vector toward target & \ref{sec:fitness} \\
\bottomrule
\end{tabular}
\end{table}

\subsection{Evolutionary Algorithms and Spatial Structure}
Conventional EAs assume \emph{panmictic} populations where any individual may mate with any other, where-as spatial EAs embed individuals in space and restrict interaction to local neighborhoods. These systems exhibit reduced global gene flow and spatial heterogeneity, with diversity maintained through isolation by distance and local competition.

Key controlling factors include dispersal distance, interaction radius, and population density, all of which influence the rate of genetic diffusion through the population and the emergence of genotypic clusters.

\subsection{HyperNEAT Controllers}
\label{sec:hyperneat}
We evolve locomotion controllers with HyperNEAT, which uses an indirect encoding: a CPPN maps neuron coordinates to connection weights, enabling regular geometric connectivity patterns. For neurons at coordinates $(x_1,y_1)$ and $(x_2,y_2)$,
\begin{equation}
w = \mathrm{CPPN}(x_1,y_1,x_2,y_2).
\end{equation}
The CPPN generates weights for a fixed substrate controller. For a gecko robot with $n$ joints, the substrate uses $n+7$ inputs (joint angles; four CPG signals; two directional inputs; bias), an $n$-unit hidden layer, and $n$ motor outputs. Connections are pruned for sparsity when $|w_{ij}|<\tau$ (typically $\tau \approx 0.2$). CPPNs are evolved with NEAT-style mutation (weight perturbation; add connection; add node) and innovation-number crossover.

\subsection{Spatial Parent Selection}
\label{sec:parent_sel}
Let $\mathbf{x}_i$ be the position of individual $i$ and $d_{ij}$ the (periodic) distance between $i$ and $j$.
\begin{itemize}
  \item \textbf{Proximity pairing:} mate with the nearest available neighbor within radius $r_{\text{pair}}$:
  \begin{equation}
  j^*=\arg\min_{j\neq i,\; j\notin P}\left\{d_{ij}\mid d_{ij}\le r_{\text{pair}}\right\},
  \end{equation}
  where $P$ is the set of already-paired individuals.
  \item \textbf{Random pairing:} randomly shuffle individuals and pair sequentially (spatially neutral control).
  \item \textbf{Mating zones:} only individuals within zones (radius $r_{\text{zone}}$ around centers $\mathbf{c}_k$) may mate:
  \begin{equation}
  \mathrm{eligible}(i,k)=\mathbb{I}\!\left[\|\mathbf{x}_i-\mathbf{c}_k\|\le r_{\text{zone}}\right],
  \end{equation}
  with proximity pairing applied within each zone. Three zone relocation strategies are available:
  \begin{enumerate}
    \item \textbf{Static:} zone centers $\mathbf{c}_k$ remain fixed throughout evolution.
    \item \textbf{Interval-based:} zones relocate uniformly at random every $n_{\text{gen}}$ generations.
    \item \textbf{Event-driven:} a zone relocates immediately after a mating event occurs within it.
  \end{enumerate}
  The event-driven strategy was designed specifically to disrupt clustering feedback loops: when mating occurs at a location, zones that remain static allow repeated mating in the same cluster, potentially causing rapid local population growth. By relocating the zone after mating, subsequent reproduction is spatially distributed, providing a responsive mechanism for population regulation.
\end{itemize}

\subsection{Movement Bias}
\label{sec:movement}
During a mating simulation window, controllers may receive optional directional inputs $\mathbf{d}_{\text{input}}$:
\begin{itemize}
  \item \textbf{Nearest-neighbor:}\;\;
  $\mathbf{d}_{\text{input}}=\dfrac{\mathbf{x}_{\text{nearest}}-\mathbf{x}_i}{\|\mathbf{x}_{\text{nearest}}-\mathbf{x}_i\|}$
  \item \textbf{Nearest-zone:}\;\;
  $\mathbf{d}_{\text{input}}=\dfrac{\mathbf{c}_{\text{nearest}}-\mathbf{x}_i}{\|\mathbf{c}_{\text{nearest}}-\mathbf{x}_i\|}$
  \item \textbf{Assigned-zone:} direction to a fixed zone assigned per individual
  \item \textbf{No bias:} $\mathbf{d}_{\text{input}}=(0,0)$
\end{itemize}

\subsection{Survival (Death) Selection}
\label{sec:death_sel}
To regulate population size and impose pressure, we implement:
\begin{itemize}
  \item \textbf{Fitness-based:} keep top $N$ by fitness.
  \item \textbf{Age-based:} keep youngest $N$ (generational replacement).
  \item \textbf{Probabilistic age:} $P(\mathrm{death}\mid i)=\min\!\left(\dfrac{\mathrm{age}_i}{\mathrm{age}_{\max}},1\right)$.
  \item \textbf{Energy-based:} individuals carry energy $E_i$ and die when $E_i\le 0$:
  \begin{equation}
  E_i(t+1)=E_i(t)-\delta E_{\text{depletion}}+\Delta E_{\text{mating}}.
  \end{equation}
  \item \textbf{Density-based:} death probability increases with local crowding:
  \begin{equation}
  \label{eq:density_death}
  P(\mathrm{death})=P_{\text{base}}+P_{\text{max}}\times\left(1-\exp\left(-\frac{\rho}{\rho_c}\right)\right),
  \end{equation}
  where local density $\rho_i = \sum_{j \neq i} \exp\left(-d_{ij}^2/2\sigma^2\right)$.
  \item \textbf{Parents die:} reproducing parents are removed, leaving offspring only.
\end{itemize}

\subsection{Periodic Boundary Conditions}
\label{sec:periodic}
To remove edge effects, we use toroidal boundaries on a world of size $W\times H$:
\begin{align}
d_{\text{periodic}}(\mathbf{x}_i,\mathbf{x}_j) &= \sqrt{d_x^2+d_y^2},\\
d_x &= \min(|x_i-x_j|,\, W-|x_i-x_j|),\\
d_y &= \min(|y_i-y_j|,\, H-|y_i-y_j|).
\end{align}

\subsection{Fitness Functions}
\label{sec:fitness}
We evaluate locomotion using either:
\begin{itemize}
  \item \textbf{Distance:}\;\;
  $f_{\text{simple}}=\|\mathbf{x}_{\text{end}}-\mathbf{x}_{\text{start}}\|$
  \item \textbf{Directional:}\;\;
  $f_{\text{directional}}=d_{\text{total}}\left(1+w_{\text{progress}}\,b_{\text{direction}}\right)$, where
  \begin{equation}
  b_{\text{direction}}=\max\!\left(0,\frac{\mathbf{v}\cdot\mathbf{u}_{\text{target}}}{\|\mathbf{v}\|}\right),\quad
  \mathbf{v}=\mathbf{x}_{\text{end}}-\mathbf{x}_{\text{start}}.
  \end{equation}
\end{itemize}

\subsection{Incubation Phase}
Optionally, we run a panmictic incubation stage for $g_{\text{incubation}}$ generations to evolve baseline locomotion before enabling spatial dynamics, then seed the spatial population with the best individuals.

\section{Methods}
This section describes the simulated platform, genotype/controller construction, the spatial evolutionary loop, and the measurements used for evaluation.

\subsection{Platform and Simulation}
\textbf{Morphology.} We use a ARIEL fixed gecko-inspired quadruped with $8$ actuated revolute joints (two per leg) in MuJoCo. Joint limits are $\pm \pi/2$ radians.

\textbf{Environment.} Individuals inhabit a flat $25\text{ m}\times 25\text{ m}$ world with periodic boundary conditions and no obstacles.

\textbf{Simulation settings.} The physics timestep is $0.002\text{ s}$. Each generation includes a $60\text{ s}$ shared-world mating simulation. Controllers are stepped every physics step and output joint position targets that are clipped to actuator limits.

\begin{figure}
    \centering
    \includegraphics[width=\linewidth]{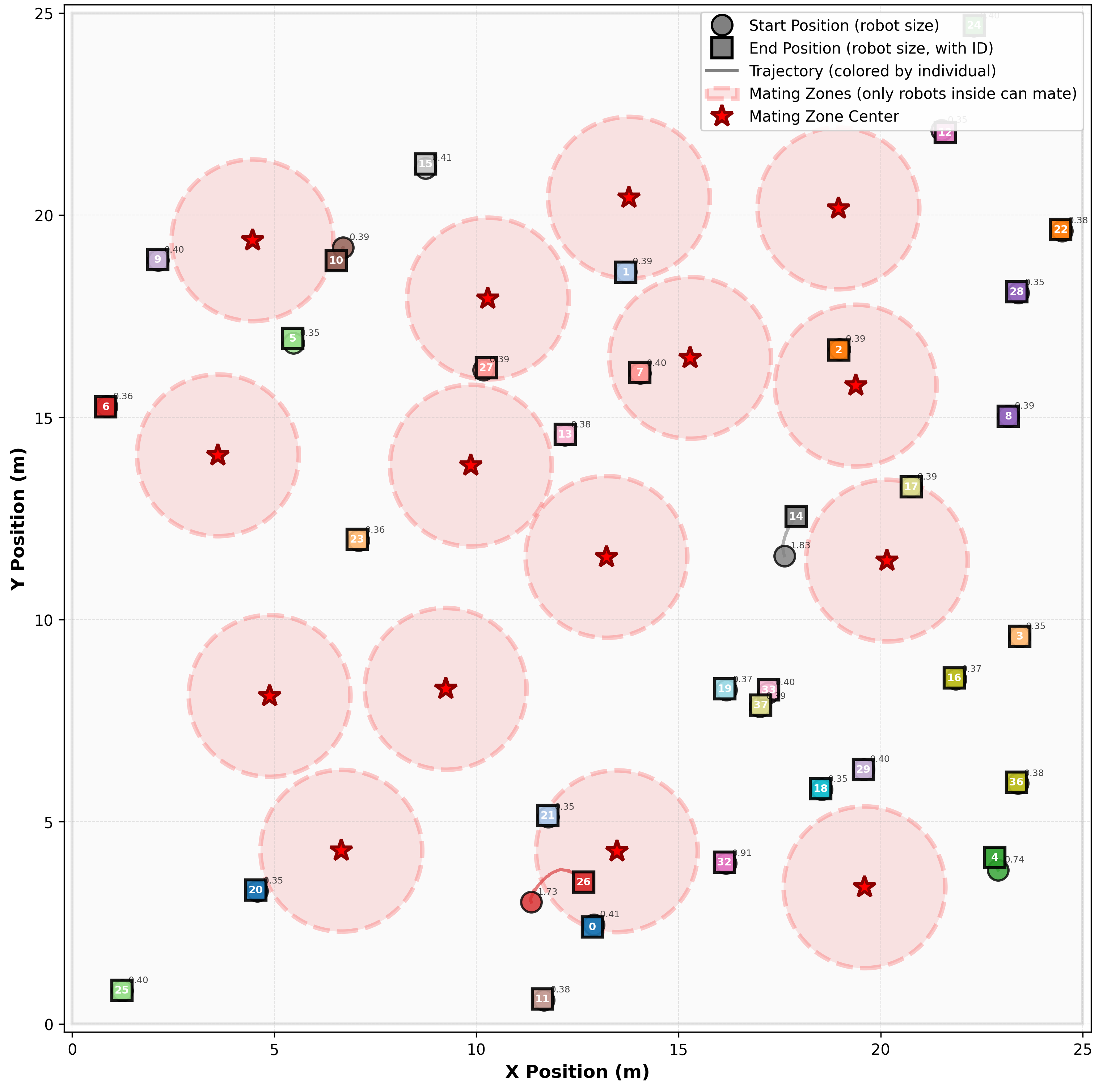}
    \caption{Example mating simulation (Generation 19). Robots (colored squares with IDs) navigate a 25m × 25m world with periodic boundaries. 
    Mating zones (dashed pink circles, radius 2m) restrict reproduction to individuals within zone boundaries. 
    Numbers show individual fitness values.}
    \label{fig:simulation_example}
\end{figure}

\textbf{Simulation Visualization.} Because the 3D simulation of robots in the environment is difficult to extract information from, an abstracted visualization is created for each generation which includes movement trajectory after each time-step, the id of specific robots, the fitness values of each, and the mating zone areas if configured. An example of this abstracted view is available in \autoref{fig:simulation_example}.

\subsection{Genotype and Genetic Operators}
\textbf{Representation.} Each individual encodes a HyperNEAT CPPN genotype with node attributes (ID, activation, layer) and connection attributes (source, target, weight, enabled flag, innovation number).

\textbf{Initialization.} CPPNs start from coordinate inputs $(x_1,y_1,x_2,y_2)$ and a single output, with a $50\%$ chance of adding $1$--$2$ hidden nodes (random activations) and broadly sampled initial weights (e.g., $\sigma=3.0$).

\textbf{Variation.} Offspring are generated by NEAT-style crossover with probability $0.9$ (aligned by innovation numbers), otherwise cloning. Mutations include weight perturbation ($p=0.8$, $\sigma_{\text{mut}}=0.5$), add-connection ($p=0.05$), and add-node ($p=0.03$).

\subsection{Controller Construction and Execution}
Each CPPN generates a fixed substrate ANN with $15$ inputs (8 joint angles, four CPG oscillators, two directional inputs, and a bias), an $8$-unit $\tanh$ hidden layer, and $8$ outputs (one per joint). At each timestep the input vector is computed from the robot state and optional movement-bias direction; the ANN output is applied to the joint controllers.

\subsection{Evolutionary Procedure}
We optionally run a short \textbf{incubation} stage: a panmictic EA for $g_{\text{incubation}}$ generations using tournament selection ($k=3$), crossover/ mutation, and elitism, to seed the initial spatial population with viable locomotion.

The main \textbf{spatial} loop runs for $G$ generations:
\begin{enumerate}
  \item \textbf{Fitness evaluation:} each individual is evaluated in isolation.
  \item \textbf{Mating simulation:} all individuals co-exist for $60$ seconds; movement bias may be enabled (nearest neighbor, nearest/assigned zone, or none).
  \item \textbf{Parent selection:} pairs are formed by proximity within radius $r_{\text{pair}}$, random pairing, or within mating zones (radius $r_{\text{zone}}$).
  \item \textbf{Reproduction:} offspring are produced (crossover or clone), mutated, and spawned near parents (radius $r_{\text{spawn}}$).
  \item \textbf{Optional dynamics:} energy is depleted each generation with configurable mating effects. Mating zones may be static, relocated at fixed intervals, or relocated via the event-driven strategy (our primary method) which moves a zone immediately after mating occurs within it to prevent persistent clustering.
  \item \textbf{Death selection:} fitness-based, age-based, probabilistic age-based, energy-based ($E\le 0$), density-based, or ``parents die'' replacement.
\end{enumerate}
Runs terminate at $G$ generations or on extinction/explosion ($N<N_{\min}$ or $N>N_{\max}$). MuJoCo objects are explicitly freed each generation; experiments are parallelized with worker recycling to limit memory growth. All distance computations use the periodic boundary conditions defined in Section~\ref{sec:periodic}, and fitness is evaluated using the functions in Section~\ref{sec:fitness}.

\subsection{Experimental Conditions and Logging}
We compare baseline pairing (random vs proximity), mating-zone sweeps (zone count, relocation strategy, movement bias), and energy dynamics (depletion and mating effects), with replicated seeds and YAML configuration files. Per generation we log population size, fitness statistics, mating counts/distances, spatial dispersion, zone occupancy (if used), energy (if used), genotypes, and movement trajectories; post-hoc analysis includes diversity metrics and genotype clustering.

\subsection{Experimental Configuration Summary}

All experiments were conducted using a fixed population size of 30 individuals initialized without incubation. Baseline experiments (random and proximity pairing with fitness-based death selection) were run for 50 generations with 10 independent runs per configuration. Spatial experiments were run for 100 generations with 45--48 independent runs per parameter configuration.

Variation operators were held constant across all experiments:

\begin{itemize}
    \item Weight mutation rate: 0.8
    \item Mutation strength ($\sigma$): 0.5
    \item Add-connection rate: 0.05
    \item Add-node rate: 0.03
    \item Crossover rate: 0.9
\end{itemize}

For spatial experiments, mating zones used a fixed radius of 2.0\,m with event-driven relocation unless otherwise specified. Offspring were spawned within a radius of 3.0\,m from parents, and pairing radius was set to 10\,m. Baseline experiments used a pairing radius of 100\,m, effectively approximating panmixia.

Energy-based selection experiments initialized each individual with energy $E_0 = 100$, applied a per-generation depletion of $\delta E = 5$, and imposed mating energy costs in $\{10, 25, 50\}$. Phase transition experiments fixed mating cost at 25 while varying the number of mating zones.

Density-based selection experiments used a Gaussian locality radius $\sigma = 3.0$\,m, base death probability $P_{\text{base}} = 0.01$, and maximum density-dependent death increment $P_{\text{max}} = 0.1$, with critical density $\rho_c \in [3.0, 7.0]$.

All experiments employed periodic boundary conditions.

\begin{table}[h]
\centering
\caption{Summary of experimental configuration across experiment families.}
\begin{tabular}{lcc}
\hline
Parameter & Baseline & Spatial Experiments \\
\hline
Population size & 30 & 30 \\
Generations & 50 & 100 \\
Independent runs & 10 & 45--48 \\
Simulation time per generation & 30\,s & 60\,s \\
Weight mutation rate & 0.8 & 0.8 \\
Mutation strength ($\sigma$) & 0.5 & 0.5 \\
Add-connection rate & 0.05 & 0.05 \\
Add-node rate & 0.03 & 0.03 \\
Crossover rate & 0.9 & 0.9 \\
Offspring spawn radius & 2.0\,m & 3.0\,m \\
Pairing radius & 100\,m & 10\,m \\
Mating zone radius & -- & 2.0\,m \\
Initial energy & -- & 100 \\
Energy depletion per generation & -- & 5 \\
\hline
\end{tabular}
\end{table}

\subsection{Evaluation Duration}

In the baseline experiments, each generation was evaluated within a 30-second simulation window. For the spatial experiments, this duration was extended to 60 seconds to allow sufficient time for encounter-driven mating dynamics to unfold. 

Consequently, direct comparisons of absolute fitness values between baseline and spatial experiments are not appropriate. However, within each experimental category, all comparisons utilized the same evaluation period, ensuring consistency of results within each group.

\section{Results}

\subsection{Baseline Experiments: Spatial vs.\ Non-Spatial Parent Selection}

To establish performance benchmarks and address \textbf{RQ1} (how spatial structure affects EA dynamics), we compared two baseline configurations using fitness-based death selection, which guarantees population stability.

Both conditions used identical parameters: 30 initial individuals, 100 generations maximum, fitness-based death selection maintaining population at target size.

\begin{table}[h!]
\centering
\begin{tabular}{@{}lcccc@{}}
\toprule
Configuration & Completed & Best & Mean Final \\
 &  & Fitness & Fitness \\
\midrule
Random Fitness-Based & 10/10 (100\%) & 1.689 & 0.412 \\
Proximity Fitness-Based & 10/10 (100\%) & 1.772 & 0.418 \\
\bottomrule
\end{tabular}
\caption{Baseline experiment results comparing spatial (proximity) and non-spatial (random) parent selection with fitness-based death selection.}
\label{tab:baseline}
\end{table}

Proximity-based pairing showed a modest 4.9\% higher peak fitness (1.772 vs 1.689) compared to random pairing, though this difference may be within stochastic variation given the limited sample size ($n=10$). Both configurations achieved 100\% completion rate, demonstrating that fitness-based death selection reliably maintains population stability regardless of parent selection method (\autoref{tab:baseline}).

\subsection{Mating Zone Experiments with Probabilistic Age Selection}

To investigate whether stochastic death selection could maintain population balance (\textbf{RQ4}), we tested probabilistic age-based death selection with mating zones across varying zone counts.

\begin{table}[h!]
\centering

\begin{tabular}{@{}lcccc@{}}
\toprule
Zones & Completed & Extinctions & Explosions & Max Gen \\
\midrule
5 & 0 (0\%) & 48 (100\%) & 0 (0\%) & 45 \\
10 & 0 (0\%) & 48 (100\%) & 0 (0\%) & 52 \\
15 & 0 (0\%) & 48 (100\%) & 0 (0\%) & 58 \\
20 & 0 (0\%) & 38 (79\%) & 10 (21\%) & 72 \\
\bottomrule
\end{tabular}
\caption{Population outcomes for probabilistic age selection across zone counts (48 runs each).}
\label{tab:probage}
\end{table}

\begin{figure}[h!]
    \centering
    \includegraphics[width=1\linewidth]{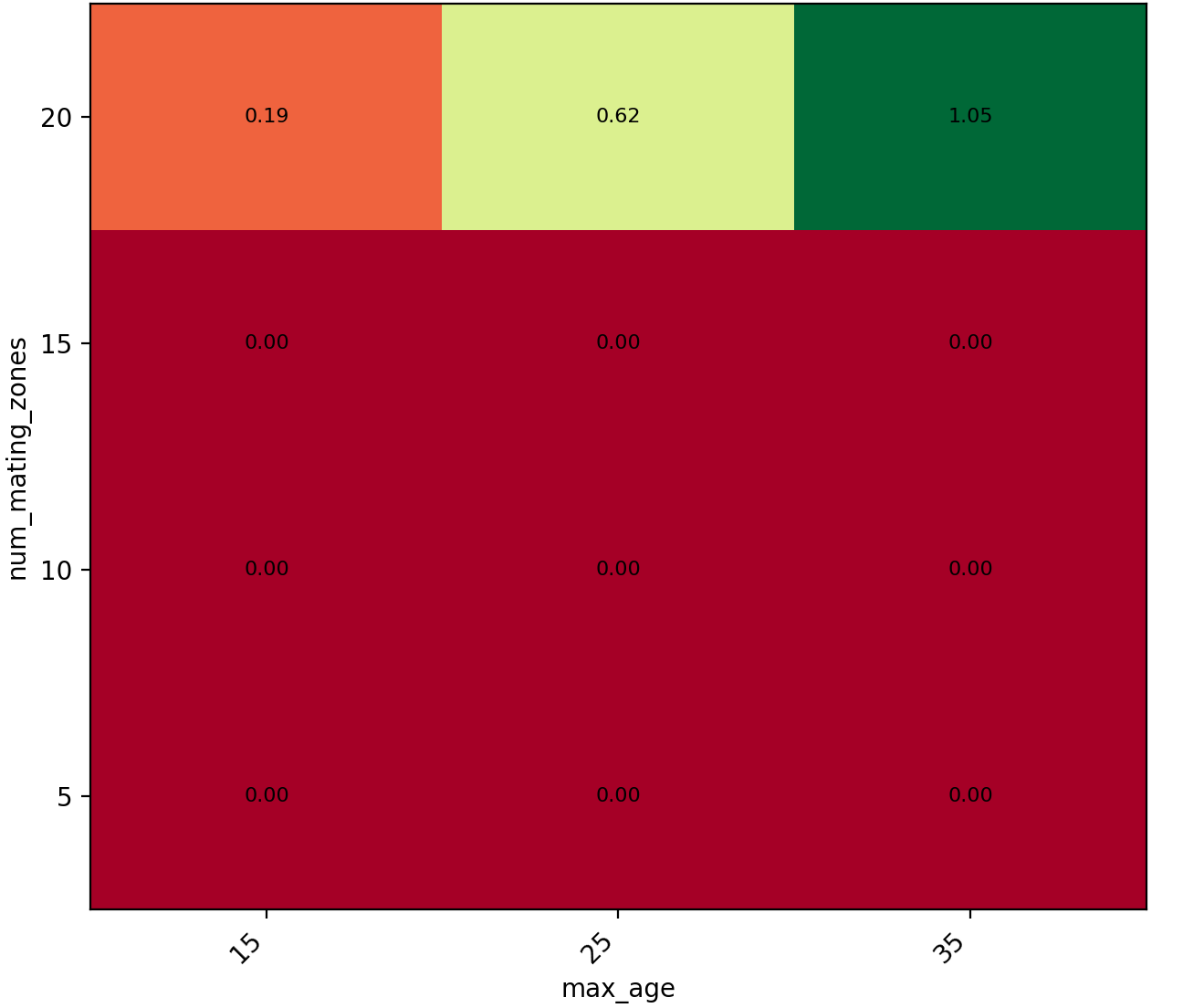}
    \caption{Final fitness averaged over 48 runs per grid search combination of number of mating zones vs the probabilistic maximum age.}
    \label{fig:grid_heatmap_final_fitness_num_mating_zones_x_max_age}
\end{figure}

In \autoref{tab:probage}, no configuration achieved stable population dynamics (0\% completion) and low zone counts (5--15) produced 100\% extinction. At 20 zones, explosions began occurring (21\%), suggesting a threshold where mating opportunities exceed mortality rate, but also where mating zone aggregate area approaches the total environmental area. In \autoref{fig:grid_heatmap_final_fitness_num_mating_zones_x_max_age}, when we begin seeing non-extinct populations at 20 zones, we also observe increasing fitness as the maximum probabilistic age rises, likely caused by decreased selection pressure, and a closeness to a global, non-zoned environment.

\subsection{Energy-Based Selection with Proximity Pairing}
We tested energy-based death selection combined with proximity pairing and nearest-neighbor movement bias.

\begin{figure}[h!]
    \centering
    \subfloat[Nearest-neighbor pairing radius $r = 1.5$m]{
        \includegraphics[width=0.9\linewidth]{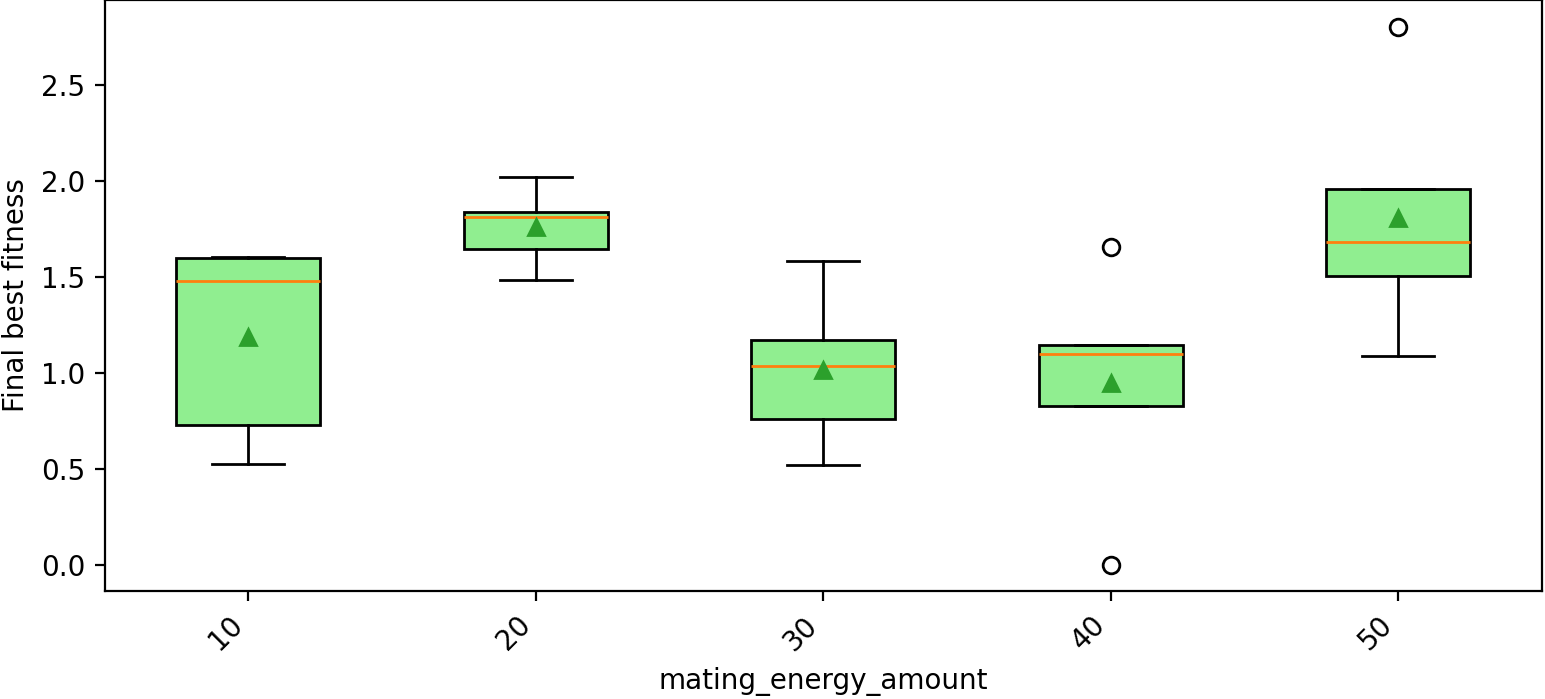}
        \label{fig:grid_conditional_fitness_radius1.5}
    }\\
    \subfloat[Nearest-neighbor pairing radius $r = 2.0$m]{
        \includegraphics[width=0.9\linewidth]{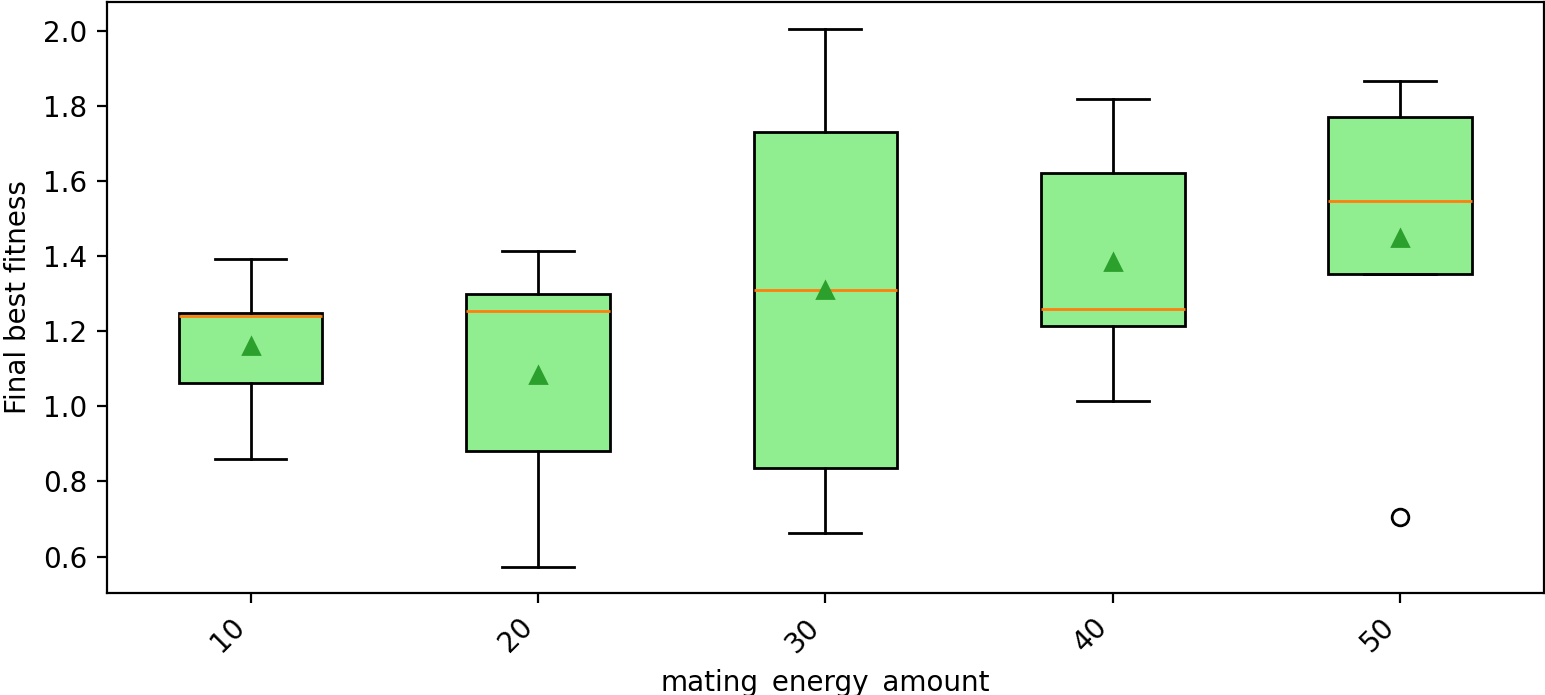}
        \label{fig:grid_conditional_fitness_radius2.0}
    }
    \caption{Final fitness averaged over 48 runs for a range of mating zone counts under different nearest-neighbor pairing radii.}
    \label{fig:grid_conditional_fitness_mating_energy_pairing_radius}
\end{figure}

100\% of runs terminated due to population explosion with fewer than $10$ generations total in any given experiment, where-in proximity pairing with nearest-neighbor movement creates a positive feedback loop in which individuals successfully chase and mate, producing offspring that also effectively chase mates within the same area. Energy depletion cannot keep pace with reproduction rate when mating is too efficient and population explosions occur too quickly to make any clear conclusions about fitness as observed in \autoref{fig:grid_conditional_fitness_radius1.5}, but expanding the radius to $2.0$m instead of $1.5$m does seem to provide a slight increase in fitness with increasing mating zones in \autoref{fig:grid_conditional_fitness_radius2.0}. This may be due to the approach towards the baseline experiment as the number of mating zones approaches the global maximum environment size, and the increase in interaction radius approaches a global interaction radius.

\subsection{Dynamic Mating Zones with Energy-Based Selection}

Our most comprehensive experiment combined event-driven mating zone relocation, assigned zone movement bias, and energy-based death selection with mating energy costs. We selected event-driven relocation as our primary mating zone strategy based on the hypothesis that immediate zone relocation after mating would disrupt spatial clustering feedback loops---a key contributor to population instability observed in preliminary trials with static zones.

\begin{figure}
    \centering
    \subfloat[Final average population]{
        \includegraphics[width=1\linewidth]{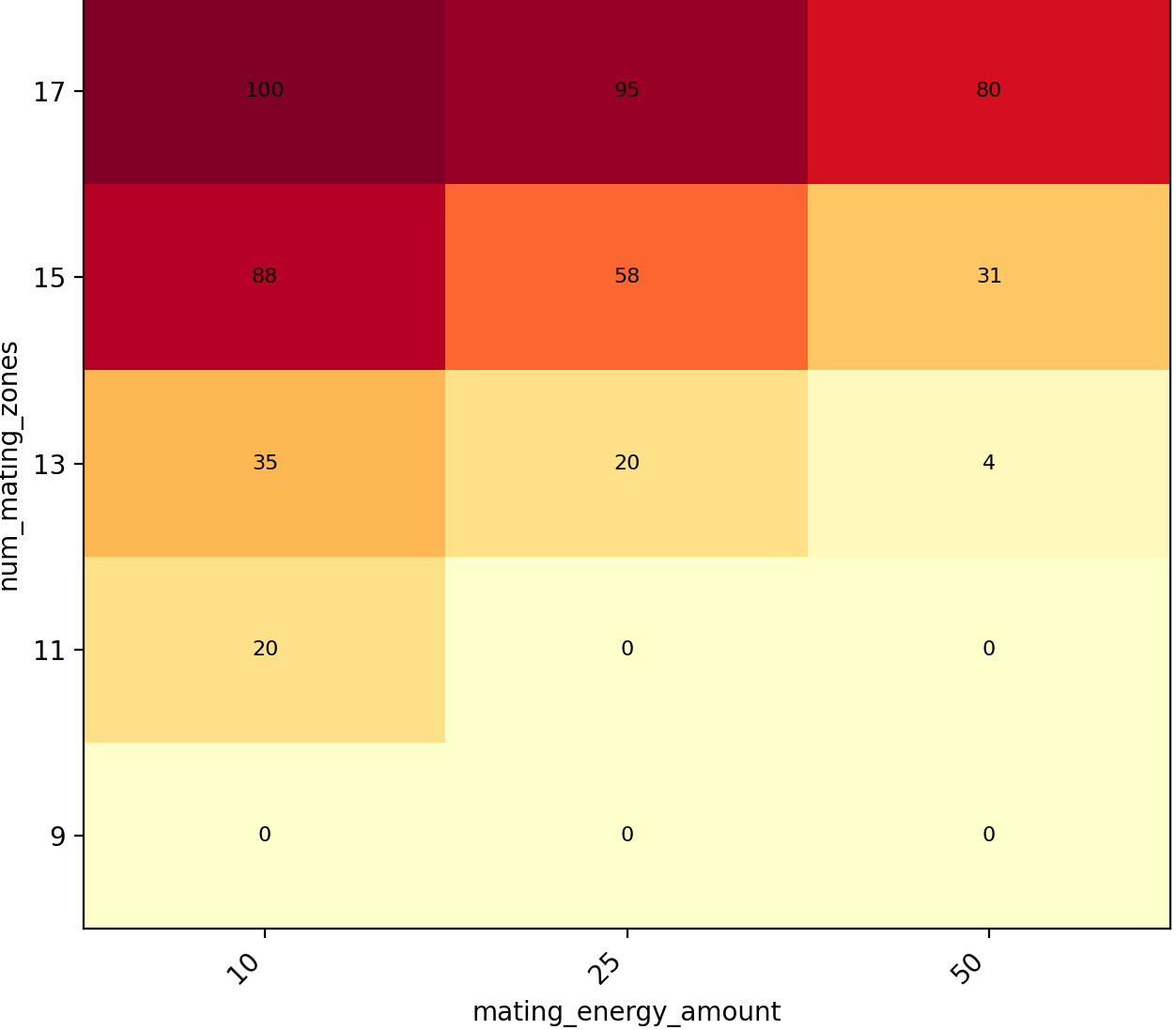}
        \label{fig:heatmap_final_population_num_mating_zones_x_mating_energy_amount}
    }\\
    \subfloat[Final best fitness]{
        \includegraphics[width=1\linewidth]{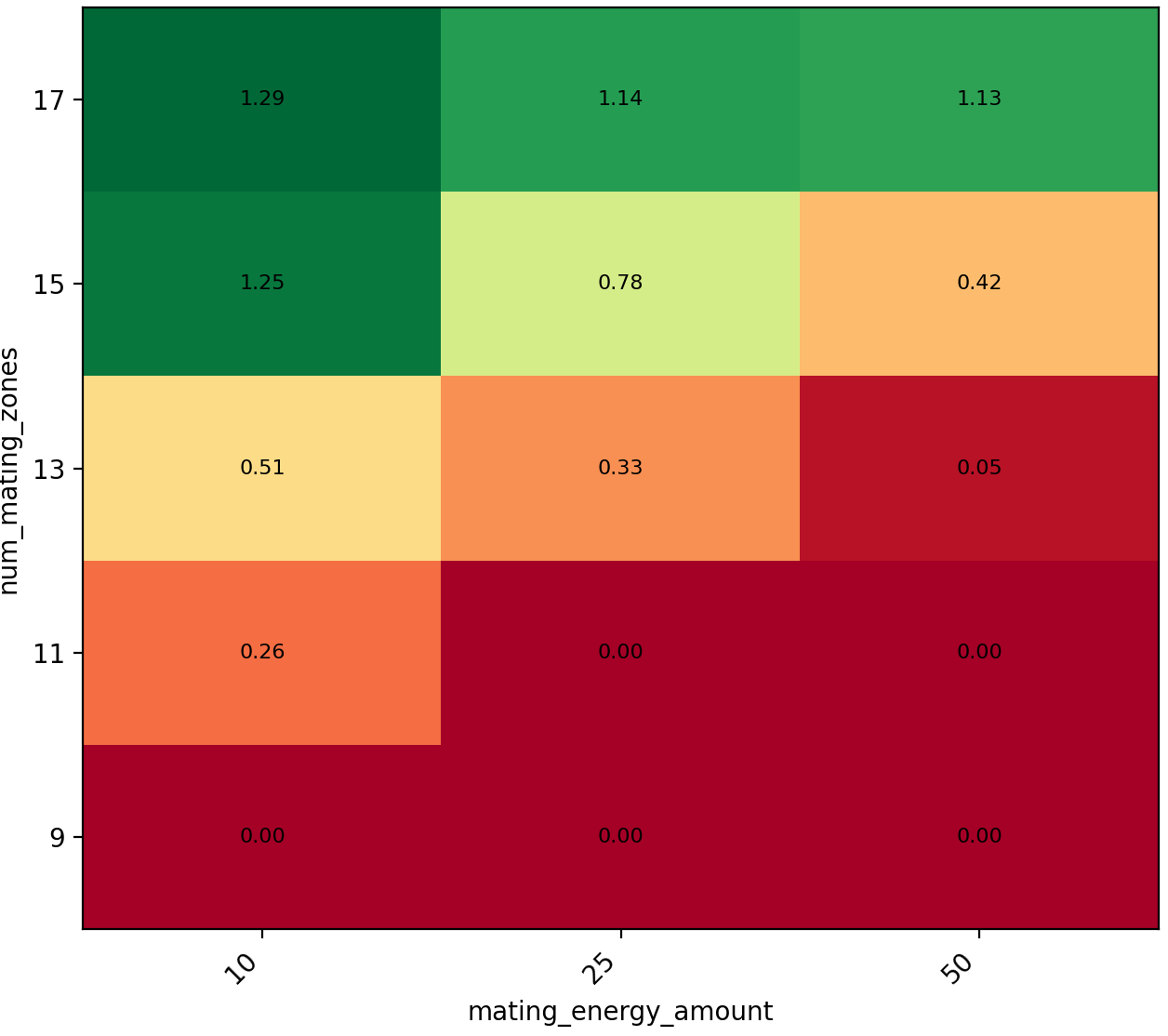}
        \label{fig:heatmap_final_fitness_num_mating_zones_x_mating_energy_amount}
    }
    \caption{Grid search over combinations of mating energy cost and mating zone counts averaged over 48 runs each.}
    \label{fig:heatmap_num_mating_zones_x_mating_energy_amount}
\end{figure}

In Figures \ref{fig:heatmap_final_population_num_mating_zones_x_mating_energy_amount} and \ref{fig:heatmap_final_fitness_num_mating_zones_x_mating_energy_amount}, we begin to observe a distinct dynamic between the number of mating zones and the mating energy cost where-in the system transitions from a state of extinction to explosion alongside a corresponding transition in resulting fitness values. Like in the previous experiments, we are able to observe the approach towards a global environmental interaction space as the number of mating zones increases, leading to both an increase in fitness as matings become more frequent, but also a push towards population explosions.

\begin{figure}[h!]
    \centering
    \includegraphics[width=1\linewidth]{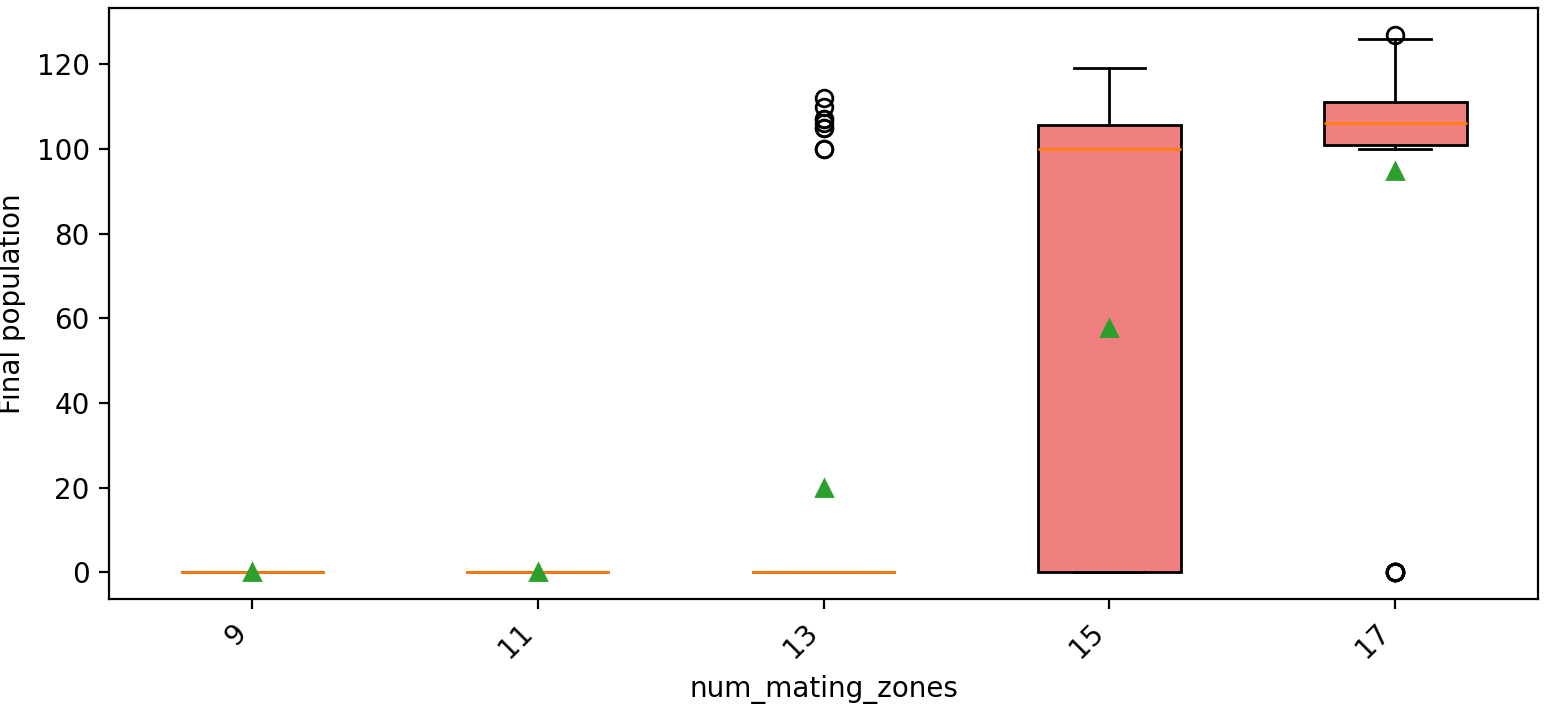}
    \caption{Final population distribution for $48$ runs with $15$ mating zones and $25$ mating energy cost.}
    \label{fig:grid_conditional_pop_num_mating_zones_given_max_age=35}
\end{figure}

What becomes most interesting is the state of transition between population explosion and population extinction where we might see stable population dynamics or non-deterministic complex system behaviors as observed in Figure \ref{fig:grid_conditional_pop_num_mating_zones_given_max_age=35}.

\begin{figure}
    \centering
    \includegraphics[width=0.9\linewidth]{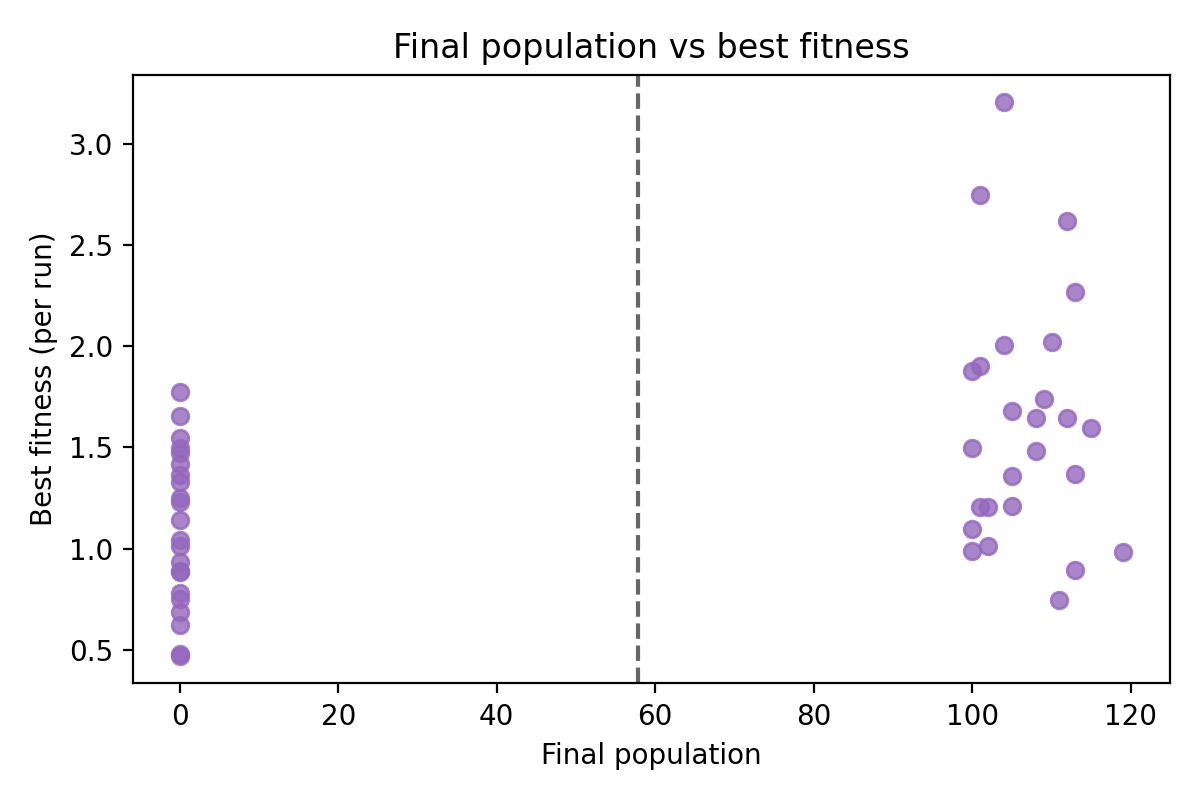}
    \caption{Final population vs Best final fitness for $48$ runs with $15$ mating zones and $25$ mating energy cost. Simulations stop and results are recorded when the population reaches $0$ or when it is greater than or equal to $100$ robots.}
    \label{fig:finalpop_vs_bestfitness}
\end{figure}

By selecting a set of grid search parameters from Figure \ref{fig:heatmap_final_population_num_mating_zones_x_mating_energy_amount} where the population average is around $50$, we can begin to observe the dynamics that exist within the system in Figure  \ref{fig:finalpop_vs_bestfitness}. This scatter plot shows how the system is not expressing any stable behavior, and much more likely expressing complex behaviors that could be expressing itself as a power-law transition around this parameter specification.

\subsection{Phase Transition Analysis}

We define an order parameter $\phi$ in \autoref{eq:phi} as the ratio between extinction and explosion in our mating energy cost experiments in \autoref{tab:order_param}.
\begin{table}[H]
\centering
\begin{tabular}{@{}cccc@{}}
\toprule
Zone Count & Order Parameter $\phi$ & Extinction Rate & Explosion Rate \\
\midrule
9 & $-1.000$ & 100\% & 0\% \\
11 & $-0.875$ & 94\% & 6\% \\
13 & $-0.625$ & 81\% & 19\% \\
15 & $+0.097$ & 45\% & 55\% \\
17 & $+0.694$ & 15\% & 85\% \\
\bottomrule
\end{tabular}
\caption{Order parameter as a function of zone count, averaged across energy cost conditions.}
\label{tab:order_param}
\end{table}
\begin{equation}
\label{eq:phi}
\phi = \frac{N_{\text{explosions}} - N_{\text{extinctions}}}{N_{\text{total runs}}}
\end{equation}

Linear interpolation between the sign change from $14$ zones to $15$ zones, the estimated critical transition point for mating zone count is $n_c \approx 14.7$, where at this critical point, the system exhibits maximum unpredictability and runs have approximately equal probability of either fate.

The system exhibits bistable dynamics rather than deterministic stability, leading to populations that do not stabilize at intermediate values but instead drift toward one of two absorbing boundaries of extinction or explosion.

\subsection{Density-Dependent Death Selection: A Negative Result}

Based on the phase transition analysis revealing decoupled reproduction and death processes, we designed a density-dependent death selection mechanism intended to couple mortality directly with spatial configuration in \autoref{eq:density_death}.

We conducted a grid search over critical density $\rho_c$ with fixed parameters chosen to create meaningful density-dependent pressure without immediate extinction:

\begin{table}[H]
\centering
\begin{tabular}{@{}lcc@{}}
\toprule
Parameter & Symbol & Value \\
\midrule
Locality radius (kernel width) & $\sigma$ & 3.0\,m \\
Base death probability & $P_{\text{base}}$ & 0.01 \\
Maximum density death probability & $P_{\text{max}}$ & 0.1 \\
Zone counts tested & --- & 13, 15 \\
Critical density range & $\rho_c$ & 3.0--7.0 \\
\bottomrule
\end{tabular}
\caption{Fixed parameters for density-based selection experiments.}
\label{tab:density_params}
\end{table}

The locality radius $\sigma = 3.0$\,m means individuals within approximately 6\,m contribute significantly to local density. The base death probability $P_{\text{base}} = 0.01$ ensures even isolated individuals face some mortality, while $P_{\text{max}} = 0.10$ caps the additional death probability from crowding---at high density ($\rho \gg \rho_c$), total death probability approaches $P_{\text{base}} + P_{\text{max}} = 0.11$.

\begin{table}[H]
\centering
\begin{tabular}{@{}cccccc@{}}
\toprule
Zones & $\rho_c$ & Completed & Extinct & Exploded & Fitness \\
\midrule
13 & 3.0 & 44 & 1 & 0 & Declining \\
13 & 5.0 & 44 & 1 & 0 & Declining \\
13 & 7.0 & 45 & 0 & 0 & Declining \\
15 & 3.0 & 43 & 1 & 1 & Declining \\
15 & 5.0 & 44 & 0 & 1 & Declining \\
15 & 7.0 & 42 & 1 & 2 & Declining \\
\bottomrule
\end{tabular}
\caption{Density-based selection results (45 runs each). Completion improved dramatically but fitness declined.}
\label{tab:density}
\end{table}

In \autoref{tab:density}, density-based selection was able to break the previously observed bistable dynamic with a 96--100\% completion rate across all experiments and parameter sets, but the goal of reducing clustering in the system of robots also saw fitness dramatically decline over generations.

\begin{figure}[H]
    \centering
    \subfloat[Aggregate population]{
        \includegraphics[width=0.9\linewidth]{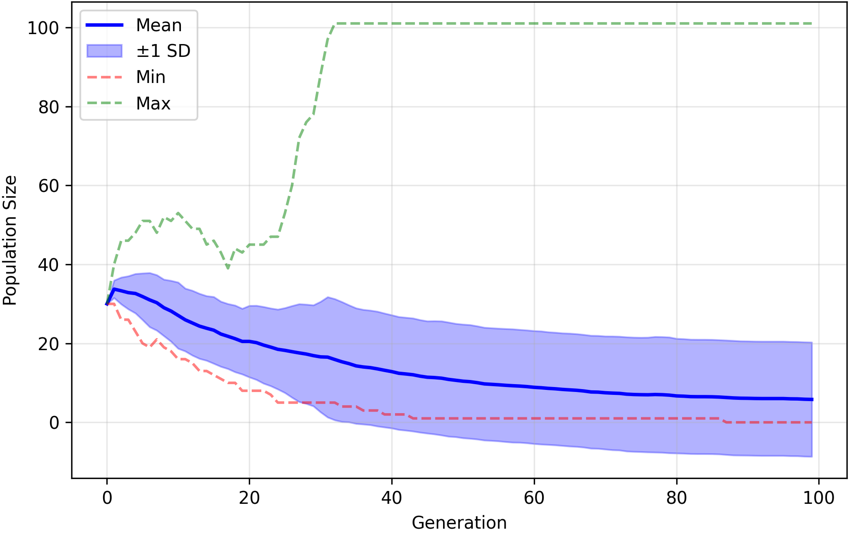}
        \label{fig:density_pop_15_3}
    }\\
    \subfloat[Aggregate fitness]{
        \includegraphics[width=0.9\linewidth]{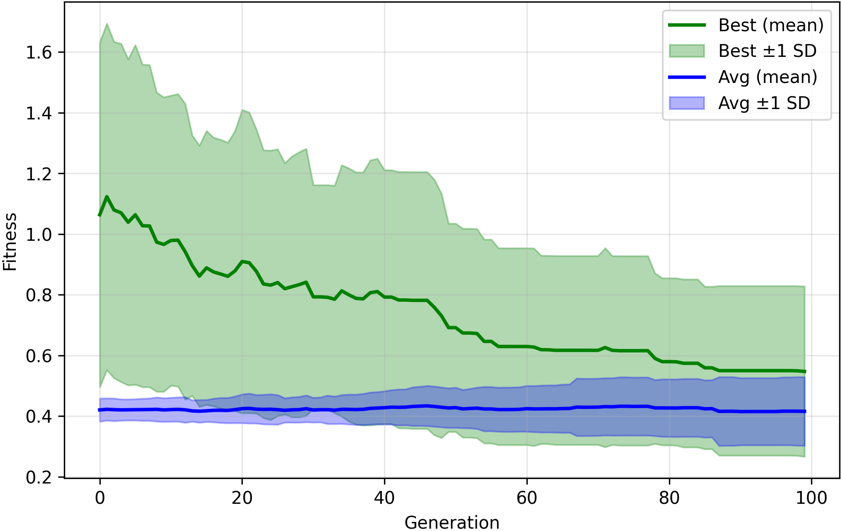}
        \label{fig:density_fit_15_3}
    }
    \caption{Aggregate results over $100$ generations for $45$ simulations with $15$ mating zones and $3.0$ critical density.}
    \label{fig:density_death_over_time}
\end{figure}



Ultimately, the density mechanism seems to create an incentive structure that punishes the exact behavior required for reproduction, with mating requiring navigation to zones that innately cause high local density, and survival requiring spatial isolation that makes robots less to mate.

The "winning strategy" under density-based selection appears to be mediocre at locomotion, where-in individuals that move effectively find mates, cluster, and die from crowding, while individuals that move poorly remain isolated, survive, but cannot reproduce anyway.


\section{Discussion}

\subsection{Research Questions}

\subsubsection*{RQ1: How does spatial structure affect EA dynamics?}

Spatial parent selection introduces a coupled relationship between reproductive success and locomotion capability which leads to bistable population dynamics when combined with stochastic death selection. Proximity-based pairing showed a modest 4.9\% higher peak fitness than random pairing, though this difference may reflect stochastic variation rather than a systematic advantage of spatial structure.

\subsubsection*{RQ2: How do different selection pressures impact adaptation?}

Fitness-Based Selection reliably maintains population stability but decouples spatial dynamics from survival and avoids fully integrating the Embodied Evolution process. Probabilistic Age Selection fails to balance reproduction rates in spatial contexts with a 0\% completion rate. Energy-Based Selection creates self-limiting dynamics in principle, but our results show energy dynamics alone cannot achieve stability due to the dynamic relationship between reproduction and death. Density-Based Selection managed to create stable populations that don't go extinct or explode, but create a perverse incentive structure with parent selection methods.

\subsubsection*{RQ3: How can parent selection be modified spatially?}

We implemented three spatial adaptations: 
\begin{enumerate}
    \item \textbf{Proximity Pairing} : individuals mate with nearest neighbors, preserving locality but providing no mating rate control
    \item \textbf{Mating Zone Constraints} : partitioning the environment creates controllable spatial structure
    \item \textbf{Movement Bias} : providing directional information transforms locomotion into mate-seeking behavior.
\end{enumerate}
Assigned zone movement with event-driven relocation proved most controllable.

\subsubsection*{RQ4: Can selection operators balance population growth?}

No tested configuration of spatial selection operators achieved stable population dynamics with positive fitness trends when using stochastic, energy-based, or density-based death selection.


Decoupled mechanisms like age and temporal energy produce bistable dynamics with no equilibrium, while positively coupled mechanisms such as death correlated with mating success cause anti-selection and fitness decline. Deterministic mechanisms such as fitness-based selection ultimately maintain stability but remove spatial effects from survival and move the model towards the abstract panmictic model.

\subsection{Critical Phenomena and Phase Transitions}

The phase transition observed when coupling dynamic mating zone movement and energy-based mating costs exhibits characteristics common to critical phenomena in complex systems, such as continuous phase transitions, where the order parameter $\phi$ varies continuously through zero, and Bistability, where two stable attractors exist with an unstable critical point between them.

The critical point (15 zones, cost=25, $\phi = 0.083$) represents maximum sensitivity to fluctuations, not stability, where each run is poised between two extreme attractors, with stochastic perturbations determining their fates.

\subsection{Implications for Spatial EA Design}

The answer to RQ4 sees that stable population balance requires mechanisms that do not directly penalize mating behavior while also allowing for parent selection methods that are spatial and dynamic. Potential approaches include:
\begin{itemize}
  \item Limiting reproduction rate rather than killing successful reproducers
  \item Preventing low-fitness individuals from mating rather than killing high-fitness clusters
  \item Using adaptive control systems that adjust parameters based on population state
\end{itemize}

\subsection{Limitations}

Our simulation fixes both morphology and environment; spatial effects may differ with terrain variation or morphological co-evolution. Compute constraints limited parameter exploration despite using the Snellius supercomputer, and the 100-generation maximum may miss longer-term dynamics in runs that achieved stable populations.

\section{Conclusion}

We presented a Spatially Embedded Evolutionary Algorithm framework for investigating how spatial structure affects evolutionary dynamics in embodied robotic systems in which spatial structures enhance selection; certain selection methods see population dynamic phase transitions; and a trade-off exists between fitness improvements, spatial implementation, and population stability.

Proximity-based pairing showed a modest 4.9\% higher peak fitness than random pairing, though this difference may be within noise and requires further investigation to confirm whether spatial constraints meaningfully improve selective pressure.

Energy-based selection experiments revealed a continuous phase transition with critical mating zone count $n_c \approx 14.7$ in dynamic mating zone configurations, separating extinction-dominated from explosion-dominated regimes. This bistability prevents stable population maintenance through selection dynamics alone.

Our systematic investigation of death selection mechanisms revealed a fundamental constraint where decoupled mechanisms like age and temporal energy produce bistable dynamics, positively coupled mechanisms such as density-based selection create anti-selection that degrades fitness, and only deterministic fitness-based selection maintains both stability and evolutionary pressure.

The positively-coupled density selection mechanism achieved 97\% completion rates over $100$ generations, but caused systematic fitness decline by punishing the clustering behavior required for reproduction, demonstrating that naive coupling of death probability to spatial configuration is counterproductive.

Effective population control for spatial EAs likely requires mechanisms that limit reproduction rate rather than penalize successful reproducers, including possible approaches such as zone-level carrying capacity, energy-gated mating prerequisites, or adaptive parameter control that respond to population state without creating perverse incentives. It shows that simple death and parent selection operators in spatial environments cannot be defined as simply as those in the abstract EAs, requiring some means by which to balance between death and reproduction in a system where robots are incentivised to constantly improve their movement capabilities to find mates.

These findings bridge evolutionary computation with concepts from complex system studies and ecology , suggesting that spatial EAs occupy a rich dynamical landscape where parameter tuning alone cannot achieve stability and structural innovations in selection mechanism design are required.

\newpage


\bibliographystyle{plainnat}
\bibliography{references}

\begin{thebibliography}{11}
\providecommand{\natexlab}[1]{#1}
\providecommand{\url}[1]{\texttt{#1}}
\expandafter\ifx\csname urlstyle\endcsname\relax
  \providecommand{\doi}[1]{doi: #1}\else
  \providecommand{\doi}{doi: \begingroup \urlstyle{rm}\Url}\fi

\bibitem[don()]{doncieux_new_2011}
New horizons in evolutionary robotics: Extended contributions from the 2009 {EvoDeRob} workshop.
\newblock URL \url{https://link.springer.com/10.1007/978-3-642-18272-3}.

\bibitem[kro()]{kroc_simulating_2010}
Simulating complex systems by cellular automata.
\newblock URL \url{http://link.springer.com/10.1007/978-3-642-12203-3}.

\bibitem[pap()]{pappa_genetic_2023}
Genetic programming: 26th european conference, {EuroGP} 2023, held as part of {EvoStar} 2023, brno, czech republic, april 12–14, 2023, proceedings.
\newblock URL \url{https://link.springer.com/10.1007/978-3-031-29573-7}.

\bibitem[Bredeche et~al.({\natexlab{a}})Bredeche, Haasdijk, and Prieto]{bredeche_embodied_2018}
Nicolas Bredeche, Evert Haasdijk, and Abraham Prieto.
\newblock Embodied evolution in collective robotics: A review.
\newblock 5:\penalty0 12, {\natexlab{a}}.
\newblock ISSN 2296-9144.
\newblock \doi{10.3389/frobt.2018.00012}.
\newblock URL \url{http://journal.frontiersin.org/article/10.3389/frobt.2018.00012/full}.

\bibitem[Bredeche et~al.({\natexlab{b}})Bredeche, Montanier, Liu, and Winfield]{bredeche_environment-driven_2012}
Nicolas Bredeche, Jean-Marc Montanier, Wenguo Liu, and Alan~F.T. Winfield.
\newblock Environment-driven distributed evolutionary adaptation in a population of autonomous robotic agents.
\newblock 18\penalty0 (1):\penalty0 101--129, {\natexlab{b}}.
\newblock ISSN 1387-3954, 1744-5051.
\newblock \doi{10.1080/13873954.2011.601425}.
\newblock URL \url{http://www.tandfonline.com/doi/abs/10.1080/13873954.2011.601425}.

\bibitem[Chopard et~al.()Chopard, Pictet, and Tomassinp]{chopard_parallel_2000}
Bastien Chopard, Olivier Pictet, and Marco Tomassinp.
\newblock {PARALLEL} {AND} {DISTRIBUTED} {EVOLUTIONARY} {COMPUTATION} {FOR} {FINANCIAL} {APPLICATIONS}.
\newblock 15\penalty0 (1):\penalty0 15--36.
\newblock ISSN 1063-7192.
\newblock \doi{10.1080/01495730008947348}.
\newblock URL \url{http://www.tandfonline.com/doi/abs/10.1080/01495730008947348}.

\bibitem[Diependaal()]{diependaal_how_nodate}
Renske Diependaal.
\newblock How robots met their others: a story of similarity and diversity.

\bibitem[Russo et~al.(2022)Russo, Palesi, Monteleone, Patti, Ascia, and Catania]{9774747}
Enrico Russo, Maurizio Palesi, Salvatore Monteleone, Davide Patti, Giuseppe Ascia, and Vincenzo Catania.
\newblock Medea: A multi-objective evolutionary approach to dnn hardware mapping.
\newblock In \emph{2022 Design, Automation \& Test in Europe Conference \& Exhibition (DATE)}, pages 226--231, 2022.
\newblock \doi{10.23919/DATE54114.2022.9774747}.

\bibitem[Skolicki()]{skolicki_analysis_2005}
Zbigniew Skolicki.
\newblock An analysis of island models in evolutionary computation.
\newblock In \emph{Proceedings of the 7th annual workshop on Genetic and evolutionary computation}, pages 386--389. {ACM}.
\newblock ISBN 978-1-4503-7800-0.
\newblock \doi{10.1145/1102256.1102343}.
\newblock URL \url{https://dl.acm.org/doi/10.1145/1102256.1102343}.

\bibitem[Sudholt()]{kacprzyk_parallel_2015}
Dirk Sudholt.
\newblock Parallel evolutionary algorithms.
\newblock In Janusz Kacprzyk and Witold Pedrycz, editors, \emph{Springer Handbook of Computational Intelligence}, pages 929--959. Springer Berlin Heidelberg.
\newblock ISBN 978-3-662-43504-5 978-3-662-43505-2.
\newblock \doi{10.1007/978-3-662-43505-2_46}.
\newblock URL \url{http://link.springer.com/10.1007/978-3-662-43505-2_46}.

\bibitem[Yao et~al.()Yao, Bullinaria, Burke, Kabán, Lozano, Merelo-Guervós, Rowe, Schwefel, Smith, and Tino]{yao_parallel_2004}
Xin Yao, John~A. Bullinaria, Edmund~K. Burke, Ata Kabán, José~A. Lozano, Juan~J. Merelo-Guervós, Jonathan Rowe, Hans-Paul Schwefel, James~E. Smith, and Peter Tino.
\newblock \emph{Parallel Problem Solving from Nature - {PPSN} {VIII}: 8th International Conference, Birmingham, {UK}, September 18-22, 2004, Proceedings}.
\newblock Number 3242 in Lecture Notes in Computer Science. Springer Berlin Heidelberg.
\newblock ISBN 978-3-540-23092-2.
\newblock \doi{10.1007/b100601}.

\end{thebibliography}

\end{document}